\documentclass[11pt,a4paper]{article}
\usepackage[hyperref]{acl 2019}

\usepackage{times}
\usepackage{latexsym}
\usepackage[noabbrev,capitalize]{cleveref}
\usepackage[normalem]{ulem}

\usepackage{url}

\usepackage{standalone}
\usepackage{tikz}
\usepackage{pgfplots}
\pgfplotsset{compat=1.14}
\usepackage{booktabs}

\pgfplotscreateplotcyclelist{tb}{
semithick,orange,mark=+\\%
semithick,blue,mark=+\\%
semithick,red,mark=+\\%
semithick,cyan,mark=+\\%
}

\aclfinalcopy 

\title{Unsupervised Pretraining for Neural Machine Translation \\
       Using Elastic Weight Consolidation}

\author{Du\v{s}an Vari\v{s} \\
  Charles University, \\
  Faculty of Mathematics and Physics \\ 
  Malostransk\'{e} n\'{a}m\v{e}st\'{i} 25 \\
  118 00 Prague, Czech Republic \\
  {\tt varis@ufal.mff.cuni.cz} \\\And
  Ond\v{r}ej Bojar \\
  Charles University, \\
  Faculty of Mathematics and Physics \\ 
  Malostransk\'{e} n\'{a}m\v{e}st\'{i} 25 \\
  118 00 Prague, Czech Republic \\
  {\tt bojar@ufal.mff.cuni.cz} \\}
\date{}

\begin{document}

\maketitle
\begin{abstract}
  This work presents our ongoing research of unsupervised pretraining in neural machine translation (NMT). In our method, we initialize the weights of the encoder and decoder with two language models that are trained with monolingual data and then fine-tune the model on parallel data using Elastic Weight Consolidation (EWC) to avoid forgetting of the original language modeling tasks.
  We compare the regularization by EWC with the previous work that focuses on regularization by language modeling objectives.
  
  The positive result is that using EWC with the decoder achieves BLEU scores similar to the previous work. However,
  the model converges 2-3 times faster and does not require the original unlabeled training data during the fine-tuning stage.
  
  In contrast, the regularization using EWC is less effective if the original and new tasks are not closely related. We show that initializing the bidirectional NMT encoder with a left-to-right language model and forcing the model to remember the original left-to-right language modeling task limits the learning capacity of the encoder for the whole bidirectional context.

\end{abstract}

\section{Introduction}

Neural machine translation (NMT) using sequence to sequence architectures \cite{sutskever2014sequence, bahdanau2015neural, vaswani2017attention} has become the dominant approach to automatic machine translation.
While being able to approach human-level performance \cite{popel2018wmttransformer}, it still requires a huge amount of parallel data, otherwise it can easily overfit. Such data, however, might not always be available. At the same time, it is generally much easier to gather large amounts of monolingual data, and therefore, it is interesting to find ways of making use of such data. The simplest strategy is to use backtranslation \cite{sennrich2016backtranslation}, 
but it can be rather costly since it requires training a model in the opposite translation direction and then translating the monolingual corpus.

It was suggested by \citet{lake2017machines} that during the development of a general human-like AI system, one of the desired characteristics of such a system is the ability to learn in a continuous manner using previously learned tasks as building blocks for mastering new, more complex tasks.
Until recently, continuous learning of neural networks was problematic, among others, due to the catastrophic forgetting \cite{mccloskey1989catastrophic}. Several methods were proposed \cite{li2016learning, aljundi2017expertgate, zenke2017synaptic}, however,
they mainly focus only on adapting the whole network (not just its parts) to new tasks
while maintaining good performance on the previously learned tasks.

In this work, we present an unsupervised pretraining method for NMT models using Elastic Weight Consolidation \cite{kirkpatrick2017catastrophic}. First, we initialize both encoder and decoder with source and target language models respectively. Then, we fine-tune the NMT model using the parallel data. To prevent the encoder and decoder from forgetting the original language modeling (LM) task, we regularize their weights individually using Elastic Weight Consolidation based on their importance to that task. Our hypothesis is the following: by forcing the network to remember the original LM tasks we can reduce overfitting of the NMT model on the limited parallel data.

We also provide a comparison of our approach with the method proposed by \citet{ramachandran2017pretraining}. They also suggest initialization of the encoder and decoder with a language model. However, during the fine-tuning phase they use the original language modeling objectives as an additional training loss in place of model regularization. Their approach has two main drawbacks: first, during the fine-tuning phase, they still require the original monolingual data which might not be available anymore in a life-long learning scenario. Second, they need to compute both machine translation and language modeling losses which increases the number of operations performed during the update slowing down the fine-tuning process.
Our proposed method addresses both problems: it requires only a small held-out set to estimate the EWC regularization term and converges 2-3 times faster than the previous method.\footnote{The speedup is with regard to the wall-clock time. In our experiments both EWC and the LM-objective methods require similar number of training examples to converge.}





\section{Related Work}

Several other approaches
towards exploiting the available monolingual data for NMT have been previously proposed.

Currently, the most common method is creating synthetic parallel data by backtranslating the target language monolingual corpora using machine translation \cite{sennrich2016backtranslation}. While being consistently beneficial, this method requires a pretrained model to prepare the backtranslations. Additionally, \citet{ramachandran2017pretraining} showed that the unsupervised pretraining approach reaches at least similar performance to the backtranslation approach.

Recently, \citet{lample2019cross} suggested using a single cross-lingual language model trained on multiple monolingual corpora as an initialization for various NLP tasks, including machine translation. While our work focuses strictly on a monolingual language model pretraining, we believe that our work can further benefit from using cross-lingual language models. 

Another possible approach is to introduce an additional
reordering \cite{zhang2016sourcemono} or de-noising objectives, the latter being recently employed in the unsupervised NMT scenarios \cite{artetxe2018unsup, lample2017unsup}. These approaches try to force the NMT model to learn useful features by presenting it with either shuffled or noisy sentences teaching it to reconstruct the original input.

Furthermore, \citet{khayrallah-etal-2018-regularized} show how to prevent catastrophic forgeting during domain adaptation scenarios. They fine-tune the general-domain NMT model using in-domain data
adding an additional cross-entropy objective to restrict the distribution of the fine-tuned model to be similar to the distribution of the original general-domain model.


\section{Elastic Weight Consolidation}

Elastic Weight Consolidation \cite{kirkpatrick2017catastrophic} is a simple, statistically motivated method for selective regularization of neural network parameters. It was proposed to counteract catastrophic forgetting in neural networks during a life-long continuous training. The previous work described the method in the context of adapting the whole network for each new task. In this section, we show that EWC can be also used to preserve only parts of the network that were relevant for the previous task, thus being potentially useful for compositional learning.

To justify the choice of the parameter constraints, \citet{kirkpatrick2017catastrophic} approach the neural network training as a Bayesian inference problem. To put it into the context of NMT, we would like to find the most probable network parameters $\theta$, given a parallel data $D_{mt}$ and monolingual data $D_{src}$ and $D_{tgt}$ for source and target languages, respectively:

\small
\begin{equation}\label{eqn:ewc-basic}
p(\theta|D_{mt}\cup{}D_{src}\cup{}D_{tgt}) = \frac{p(D_{mt}|\theta) p(\theta|D_{src}\cup{}D_{tgt})} {p(D_{mt})}
\end{equation}
\normalsize

Equation~\ref{eqn:ewc-basic} holds, assuming datasets $D_{mt}$, $D_{src}$ and $D_{tgt}$ being mutually exclusive. 
The probability $p(D_{mt}|\theta)$ is the negative of the MT loss function and $p(\theta|D_{src}\cup{}D_{tgt})$ is the result of the unsupervised pretraining. We can assume that during the unsupervised pretraining, the parameters $\theta_{src}$ of the encoder are independent of the parameters $\theta_{tgt}$ of the decoder. Furthermore, we assume that the parameters of the encoder are independent of the target-side monolingual data and the parameters of the decoder are independent of the source-side monolingual data. 
Given these assumptions, we can express the posterior probability $p(\theta|D_{src}\cup{}D_{tgt})$ in the following way:

\begin{equation}\label{eqn:ewc-lm}
p(\theta|D_{src}\cup{}D_{tgt}) = p(\theta_{src}|D_{src}) p(\theta_{tgt}|D_{tgt})
\end{equation}

Probabilities $p(\theta_{src}|D_{src})$ and $p(\theta_{tgt}|D_{tgt})$ are given by the pretrained source and target language models respectively. The true posterior probabilities given by the language models are intractable during fine-tuning, however, similarly to the work of \citet{kirkpatrick2017catastrophic}, we can estimate $p(\theta_{src}|D_{src})$ as Gaussian distribution using Laplace approximation \cite{MacKay1992laplace}, with mean given by the pretrained parameters $\theta_{src}$ and variance given by a diagonal of the Fisher information matrix $F_{src}$. Then, we can add the following regularization term to our loss function:

\begin{equation}\label{eqn:ewc-term}
L_{ewc-src}(\theta) = \sum_{i, \theta_{i}\subset \theta_{src}} \frac{\lambda}{2}F_{src, i}(\theta_{i} - \theta^{\star}_{src, i})^2
\end{equation}

The model parameters not present during the language model pretraining are ignored by the regularization term. Analogically, the same can be applied for the target-side posterior probability $p(\theta_{tgt}|D_{tgt})$ giving a target-side regularization term $L_{ewc-tgt}$.



In the following section, we show that these regularization terms can be useful in a low-resource machine translation scenario.
Since we do not necessarily need to preserve the knowledge of the original language modeling tasks, we focus on using them only as prior knowledge to prevent overfitting during the fine-tuning.





\section{Experiments}


In this section, we present the results of our experiments with EWC regularization and compare them with the previously proposed regularization by language modeling objectives.

\subsection{Model Description}

In all experiments, we use the self-attentive Transformer network \cite{vaswani2017attention} because it is the current state-of-the-art NMT architecture, providing us with a strong baseline. In general, it follows the standard encoder-decoder paradigm, with encoder creating hidden representations of the input tokens based on their surrounding context and decoder generating the output tokens autoregressively while attending to the source sentence token representations and tokens it generated in the previous decoding steps.\footnote{For more details about the architecture, see the original paper.}

We use Transformer with 6 layers in both encoder and decoder. We set the dimension of the hidden states to 512 and the dimension of the feedforward layer to 2048. We use multi-head attention with 16 attention heads. To simplify the pretraining process, we use a separate vocabulary for source and target languages, each containing around 32k subwords. We use separate embeddings in the encoder and decoder. In the decoder, we tie the embeddings with the output softmax layer \cite{press-wolf-2017-using}. During both pretraining and fine tuning, we use Adam optimizer \cite{kingma2014adam} and gradient clipping. We set the initial learning rate to 3.1, use a linear warm-up for 33500 training steps and then decay the learning rate exponentially. We set the training batch size to a maximum of 2048 tokens per batch together with sentence bucketing for more efficient training. We set dropout to 0.1. During the final evaluation, we use beam search decoding with beam size of 8 and length normalization set to 1.0.

When pretraining the encoder and decoder, we use identical network parameters. 
We train each language model to maximize the probability of each word in a sentence using its leftward context.
To pretrain the decoder, we use the decoder architecture from Transformer with encoder-attention sub-layer removed due to the lack of source sentences. Later, we initialize the NMT decoder with the language model weights and the encoder-attention weights by a normal distribution. We reset all training-related variables (learning rate, Adam moments) during the NMT initialization.

For simplicity, we apply the same approach for the encoder pretraining. In \cref{sec:dataeval}, we discuss the drawbacks of our encoder pretraining and suggest possible improvements. In all experiments, we set the weight $\lambda$ of each EWC regularization term to 0.02.


\begin{table}
\begin{center}
\begin{tabular}{l|c|cccc}
\toprule
&  & SRC & TGT & ALL \\ 
\midrule
Baseline & 15.68 & -- & -- & -- \\
Backtrans. & 19.65 & -- & -- & -- \\
\midrule
LM best & -- & 13.96 & 15.56 & 16.83 \\ 
EWC best & -- & 10.77 & \textbf{15.91} & 14.10 \\
\midrule
LM ens. & -- & 15.16 & 16.60 & 17.14 \\
EWC ens. & -- & 10.73 & \textbf{16.63} & 14.66 \\
\bottomrule
\end{tabular}
\end{center}

\caption{Comparison of the previous work (LM) with the proposed method (EWC). We compared models with only pretrained encoder (SRC), pretrained decoder (TGT) and both (ALL). All pretrained language models contained 3 layers. We compared both single best models and ensemble (using checkpoint averaging) of 4 best checkpoints. Results where the proposed method outperformed the previous work are in bold.}
\label{tab:results}
\end{table}

The model implementation is available in 
Neural Monkey\footnote{\url{https://github.com/ufal/neuralmonkey}} \cite{nmonkey2017}
framework for sequence-to-sequence modeling.

\subsection{Dataset and Evaluation}
\label{sec:dataeval}

In our experiments, we focused on the low-resource Basque-to-English machine translation task featured in IWSLT 2018.\footnote{\url{https://sites.google.com/site/iwsltevaluation2018/TED-tasks}}
We used the parallel data provided by IWSLT organizers, consisting of 5,600 in-domain sentence pairs (TED Talks) and around 940,000 general-domain sentence pairs. During pretraining, we used Basque Wikipedia for source language model and NewsCommentary 2015 provided by WMT\footnote{\url{http://www.statmt.org/wmt18/translation-task.html}} for target language model. Both corpora contain close to 3 million sentences. We used UDPipe\footnote{\url{http://ufal.mff.cuni.cz/udpipe}} \cite{udpipe2017} to split the monolingual data to sentences and SentencePiece\footnote{\url{https://github.com/google/sentencepiece}} to prepare the subword tokenization. We used the subword models trained on the monolingual data to preprocess the parallel data.

\begin{figure}
    \centering
    \def\verticalstep{0.8}

\begin{tikzpicture}[]
  \begin{axis}[cycle list name=tb, 
               grid=both,
               grid style={solid,gray!30!white},
               axis lines=middle,
               xlabel={Number of Sentences},
               ylabel={BLEU},
               x label style={at={(axis description cs:0.5,-0.1)},anchor=north},
               y label style={at={(axis description cs:-0.1,.5)},rotate=90,anchor=south},
               legend pos=south east]
    \addplot table [x=Step, y=Value, col sep=comma] {./data/run_out.mt.lm-init.src.ewc_reg.eu-en.depth-2-tag-val_target_BLEU-4.csv};
    \addplot table [x=Step, y=Value, col sep=comma] {./data/run_out.mt.lm-init.src.ewc_reg.eu-en.depth-4-tag-val_target_BLEU-4.csv};
    \addplot table [x=Step, y=Value, col sep=comma] {./data/run_out.mt.lm-init.src.ewc_reg.eu-en.depth-6-tag-val_target_BLEU-4.csv};
    \addplot table [x=Step, y=Value, col sep=comma] {./data/run_out.mt.tformer.eu-en.baseline-tag-val_target_BLEU-4.csv};
    \addlegendentry{depth-2}
    \addlegendentry{depth-4}
    \addlegendentry{depth-6}
    \addlegendentry{no pretraining}
  \end{axis}
\end{tikzpicture}
    \caption{Performance of MT models where only the encoder was initialized by the language model of varying depths and then regularized by EWC. We include the performance of the MT system that was not pretrained for comparison.}
    \label{fig:degrad}
\end{figure}
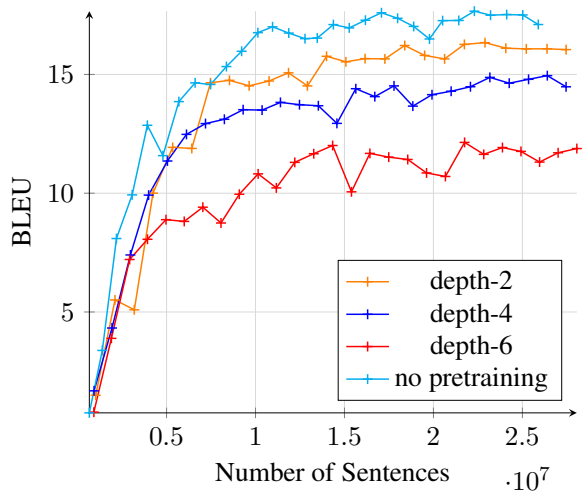

During training, we used development data provided by IWSLT 2018 organizers which contains 1,140 parallel sentences.
To approximate the Fisher Information Matrix diagonal of the pretrained Basque and English language models, we used the respective parts of the IWSLT validation set. For final evaluation, we used the IWSLT 2018 test data consisting of 1051 sentence pairs.

Table~\ref{tab:results} compares the performance of the models fine-tuned using the LM objective regularization and the EWC regularization. First, we can see that using EWC when only the decoder was pretrained slightly outperforms the previous work. On the other hand, our method fails when used in combination with the encoder initialization by the source language model. The reason might be a difference between the original LM task that is trained in a left-to-right autoregressive fashion while the strength of the Transformer encoder is in modelling of the whole left-and-right context for each source token. The learning capacity of the encoder is therefore restricted by forcing it to remember a task that is not so closely related to the sentence encoding in Transformer NMT. \cref{fig:degrad} supports our claim: the deeper the pretrained language model and therefore more layers regularized by EWC, the lower the performance of the fine-tuned NMT system.
We think that this behaviour can be mitigated by initializing the encoder with a language model that considers the whole bidirectional context, e.g. a recently introduced BERT encoder \cite{devlin2018bert}. We leave this for our future work.

\begin{figure}
    \centering
    \def\verticalstep{0.8}

\begin{tikzpicture}[]
  \begin{axis}[cycle list name=tb, 
               grid=both,
               grid style={solid,gray!30!white},
               axis lines=middle,
               xlabel={Training time (h)},
               ylabel={Perplexity},
               ymin=4.8,
               x label style={
                    at={(axis description cs:0.5,-0.1)},
                    anchor=north},
               y label style={
                    at={(axis description cs:-0.1,.5)},
                    rotate=90,
                    anchor=south},
               yticklabel style={%
                    /pgf/number format/.cd,
                    fixed,
                    fixed zerofill,
                    precision=1
                },
               legend pos=north east]
    \addplot table [x=Relative time, y=Value, col sep=comma] {./data/run_out.mt.lm-init.tgt.ewc_reg.eu-en.depth-3-tag-val_target_ppl_perplexity.csv};
    \addplot table [x=Relative time, y=Value, col sep=comma] {./data/run_out.mt.lm-init.tgt.multidata.lm_reg.eu-en.depth-3-tag-val_target_ppl_perplexity.csv};
    \addplot table [x=Relative time, y=Value, col sep=comma] {./data/run_out.mt.lm-init.tgt.no_ewc.eu-en.depth-3-tag-val_target_ppl_perplexity.csv};
    \addlegendentry{EWC-reg}
    \addlegendentry{LM-reg}
    \addlegendentry{no-reg}
  \end{axis}
\end{tikzpicture}
    \caption{Comparison of relative convergence times (measured by perplexity) of models where only the decoder was pretrained. The models were regularized using EWC, LM objective or were not using any regularization (no reg.). All models were trained on the same number of training examples ($\sim$27M sentences). All used a pretrained LM with 3 Transformer layers.}
    \label{fig:speed}
\end{figure}
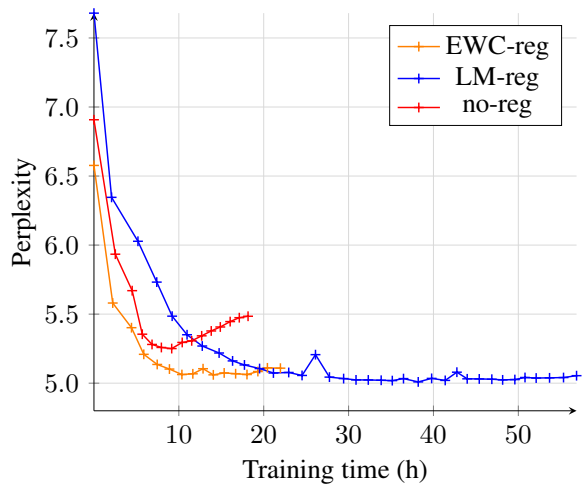

In addition to improving model performance, EWC converges much faster than the previously introduced LM regularizer. Figure~\ref{fig:speed} shows that the model fine-tuned without LM regularization converged in about 10 hours, while it took around 20-30 hours to converge when LM regularization was in place. Note, that all models converged after seeing a similar number of training examples, however, computing the LM loss for regularization introduces an additional computation overhead.
The main benefit of both EWC and LM-based regularization is apparent here, too. The non-regularized model quickly overfits.

As the comparison to the model trained on the backtranslated monolingual corpus shows, none of our regularization methods can match this simple but much more computationally demanding benchmark. 


\section{Conclusion}

We introduced our work in progress, and exploration of model regularization of NMT encoder and decoder parameters based on their importance for previously learned tasks and its application in the unsupervised pretraining scenario.
We documented that our method slightly improves the NMT performance (compared to the baseline as well as the previous work of LM-based regularization) when combined with a pretrained target language model.
We achieve this improvement at a reduced training time.

We also showed that the method is less effective if the original language modeling task used to pretrain the NMT encoder is too different from the task learned during the fine-tuning. We plan to further investigate whether we can gain improvements by using a different pretraining method for the encoder and how much this task mismatch relates to the learning capacity of the encoder.

\section*{Acknowledgments}
This work has been in part supported by the
 project no. 19-26934X (NEUREM3) of the Czech Science Foundation,
by the Charles University SVV project number 260~453 and
by the grant no. 1140218 of the Grant Agency of the Charles University.

\bibliography{naaclhlt2019}
\bibliographystyle{acl_natbib}

\end{document}